\documentclass[a4paper,conference]{IEEEtran}
\usepackage{times}
\usepackage{soul}
\usepackage{url}
\usepackage[hidelinks]{hyperref}
\usepackage[utf8]{inputenc}
\usepackage[small]{caption}
\usepackage{graphicx}
\usepackage{amssymb}
\usepackage{amsmath}
\usepackage{amsthm}
\usepackage{booktabs}
\usepackage{algorithm}
\usepackage{algorithmic}
\usepackage{colortbl}
\usepackage{cite}

\usepackage{algorithmic}
\usepackage{array}
\ifCLASSOPTIONcompsoc
 \usepackage[caption=false,font=normalsize,labelfont=sf,textfont=sf]{subfig}
\else
 \usepackage[caption=false,font=footnotesize]{subfig}
\fi
\usepackage{fixltx2e}
\usepackage{url}

\def\ie{{\em i.e.}}
\def\eg{{\em e.g.}}

\def\etal{{\em et al.}}

\begin{document}
%
\title{MFPP: Morphological Fragmental Perturbation Pyramid for Black-Box Model Explanations}

%
\author{\IEEEauthorblockN{Qing Yang\IEEEauthorrefmark{1},
Xia Zhu\IEEEauthorrefmark{2},
Jong-Kae Fwu\IEEEauthorrefmark{2},Yun Ye\IEEEauthorrefmark{1},
Ganmei You\IEEEauthorrefmark{1} and
Yuan Zhu\IEEEauthorrefmark{1}}
\IEEEauthorblockA{\IEEEauthorrefmark{1}Intel Corporation, China\\
Email: \{qing.y.yang, yun.ye, ganmei.you, yuan.y.zhu\}@intel.com}
\IEEEauthorblockA{\IEEEauthorrefmark{2}Intel Corporation, USA}
Email: \{xia.zhu,jong-kae.fwu\}@intel.com}


\maketitle

\begin{abstract}
    Deep neural networks (DNNs) have recently been applied and used in many advanced and diverse tasks, such as medical diagnosis, automatic driving, etc. Due to the lack of transparency of the deep models, DNNs are often criticized for their prediction that cannot be explainable by human. In this paper, we propose a novel Morphological Fragmental Perturbation Pyramid (MFPP) method to solve the Explainable AI problem. In particular, we focus on the black-box scheme, which can identify the input area that is responsible for the output of the DNN without having to understand the internal architecture of the DNN. In the MFPP method, we divide the input image into multi-scale fragments and randomly mask out fragments as perturbation to generate a saliency map, which indicates the significance of each pixel for the prediction result of the black box model. Compared with the existing input sampling perturbation method, the pyramid structure fragment has proved to be more effective. It can better explore the morphological information of the input image to match its semantic information, and does not need any value inside the DNN. We qualitatively and quantitatively prove that MFPP meets and exceeds the performance of state-of-the-art (SOTA) black-box interpretation method on multiple DNN models and datasets.
\end{abstract}

\ifCLASSOPTIONpeerreview
\begin{center} \bfseries EDICS Category: 3-BBND \end{center}
\fi
%
\IEEEpeerreviewmaketitle

\section{Introduction}
\label{sec:intro}
In the past decade, deep neural networks (DNN) have made breakthroughs in various AI tasks and greatly changed many fields, such as computer vision and natural language processing. However, the lack of transparency of the DNN model has led to serious concerns about the widespread deployment of machine learning technologies, especially when these DNN models are given decision-making power in critical applications such as medical diagnosis, autonomous driving, intelligent 
surveillance and financial authentication~\cite{pipenet} \etal.

\begin{figure*}[ht]
	\begin{center}
		\centering
		\includegraphics[height=11.0cm,width=18cm]{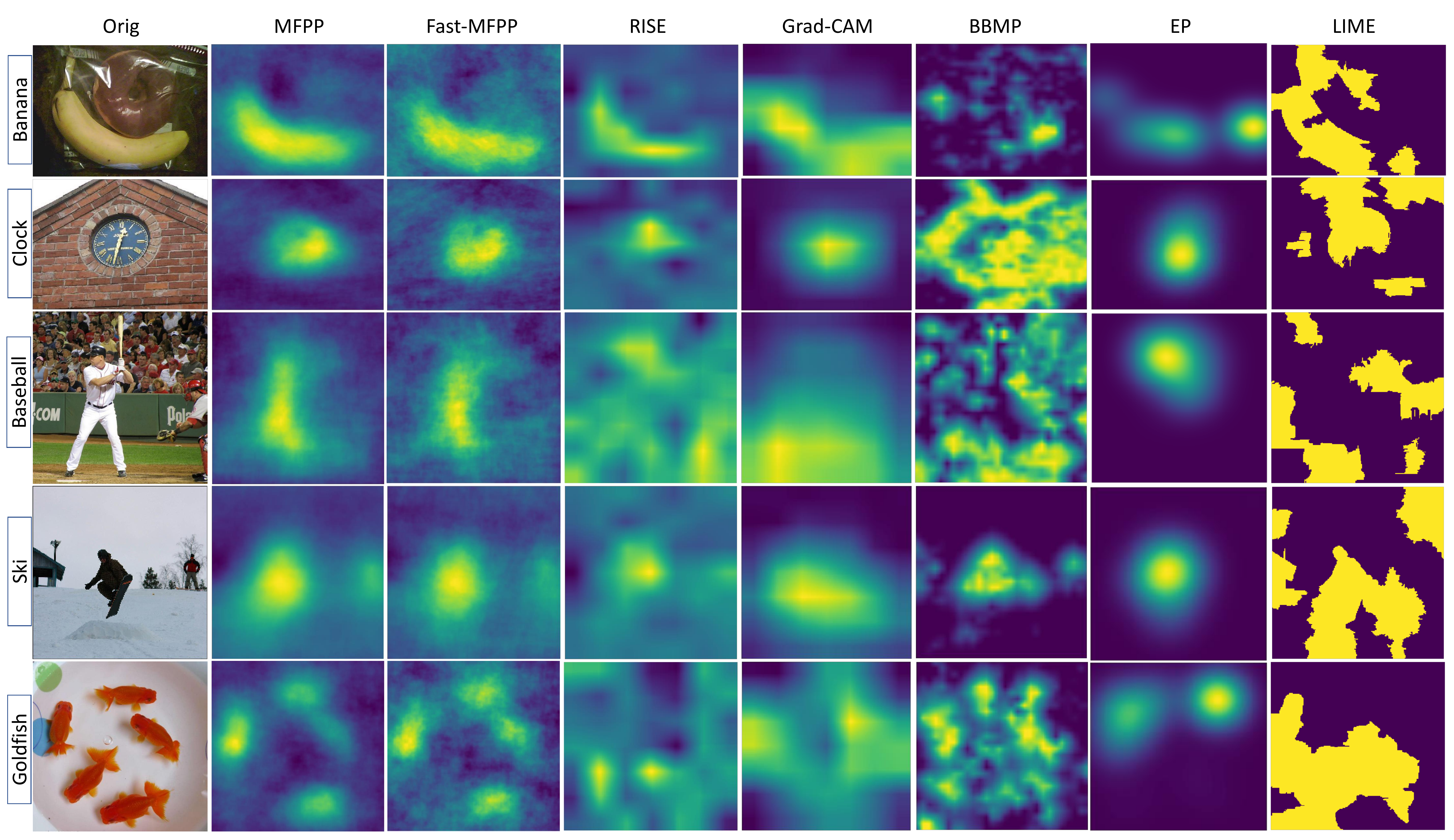}
	\end{center}
	\caption{The visualization comparison results of the proposed MFPP, the proposed Fast-MFPP, and four other methods from left to right: (EP)~\cite{ep}, RISE~\cite{rise}, BBMP~\cite{bbmp}, LIME~\cite{lime} and Grad-CAM~\cite{gradcam} on VGG16. Samples $1-4$ are from MS COCO2014~\cite{Lin_2014}. Sample $5$ is public image~\cite{goldfish}. }
\label{fig:benchmark}
\end{figure*}

Many methods for model interpretation have been proposed, but often produce unsatisfactory results. Some existing methods~\cite{Zhou_2016,gradcam,gradcampp} generate saliency maps based on intermediate information. For example, CAM~\cite{Zhou_2016}(Class Activation Mapping) requires to add or use existing global average pooling layer before the last fully-connected layer to generate saliency map; Grad-CAM~\cite{gradcam} needs weights and feature map values inside a DNN to get weighted summation. In addition to requiring DNN internal structure information, the results are with low resolution and visually coarse due to the up-scaling operations in the saliency map generation process. As shown in Figure~\ref{fig:benchmark}, Grad-CAM can locate objects such as bananas and skiers, but the generated map is larger than the object itself, which reduces its reliability and effectiveness. The black-box model methods~\cite{lime,bbmp,rise,fgvis} do not require the internal structure and values of DNNs, but there are still problems with accuracy and granularity. LIME~\cite{lime} proposes a model agnostic method with traditional machine learning ideas, and interprets the prediction of more complex models by fitting a small local linear models. However, the linear model cannot fully handle the weight of millions of data, which may lead to underfitting. Therefore, they simplify the input sampling method (from pixel mode to superpixel mode), resulting in a coarse-grained and less accurate saliency map. For objects such as banana and wall-clock in Figure~\ref{fig:benchmark}, the saliency map from LIME covers part of the target object, but also covers other regions. Its non-adaptive probability threshold value needs user's manual input. BBMP~\cite{bbmp} uses the Adam optimizer to iteratively optimize the saliency map. However, when the background is complex and deceptive, it is difficult to converge and correctly locate the target. As a result, BBMP makes a poor explanation in the first three samples of Figure~\ref{fig:benchmark}. RISE~\cite{rise} can generate a saliency map for each pixel, and it works well on rectangular objects (such as the goldfish sample in the fifth row). When non-rectangular targets and multiple tiny objects belong to the same category, grid-based sampling methods (such as RISE) will cause performance degradation. Recent work Extremal Perturbation (EP)~\cite{ep} is the best prior method, which can identify saliency maps through extreme perturbation and SGD. Our evaluation results in Sec.\ref{evaluation} show that EP is time-consuming and provides unbalanced accuracy.

In summary, the main contributions of this paper are summarized as follows:

(1) A novel Morphological Fragmentation Perturbation Pyramid (MFPP) design is proposed for black box model interpretation, which perturbs morphological fragments of different scales and makes full use of input semantic information. 
(2) The proposed MFPP is applied to the random input sampling method and significantly improve its performance in interpreting accuracy scores. 
(3) The qualitative and quantitative evaluations on multiple data sets and models are performed. It is proven that MFPP has better interpretability for black box DNN prediction, higher accuracy, and an order of magnitude faster than the latest methods.  

The rest of the paper is organized as follows: Section~\ref{related} briefly reviews related works in model explanation area. Section~\ref{method} proposes the black-box model explanation method MFPP. Section~\ref{experiment} shows the intuitive and quantitative experiments and corresponding results analysis. Finally, a summary of the paper is provided in Section~\ref{conclusion}.

\section{Related work}
\label{related}
In the past few years, many methods~\cite{Zhou_2016,gradcam,gradcampp,lime,rise,fgvis} have been proposed to explain and visualize the predictions of deep CNN classifiers, which greatly promotes the research of model interpretability and design optimization.  
Survey papers~\cite{zhang2018visual}, ~\cite{guidotti2019survey}  and ~\cite{du2019techniques} have fully summarized these methods. 
In this section, we provide some criteria to classify the previous visual interpretation methods in multiple dimensions. This can allow researchers to change their views to thoroughly understand the differences between each category.
\begin{figure*}[ht]
	\begin{center}
		\centering
		\includegraphics[height=7.5cm,width=16cm]{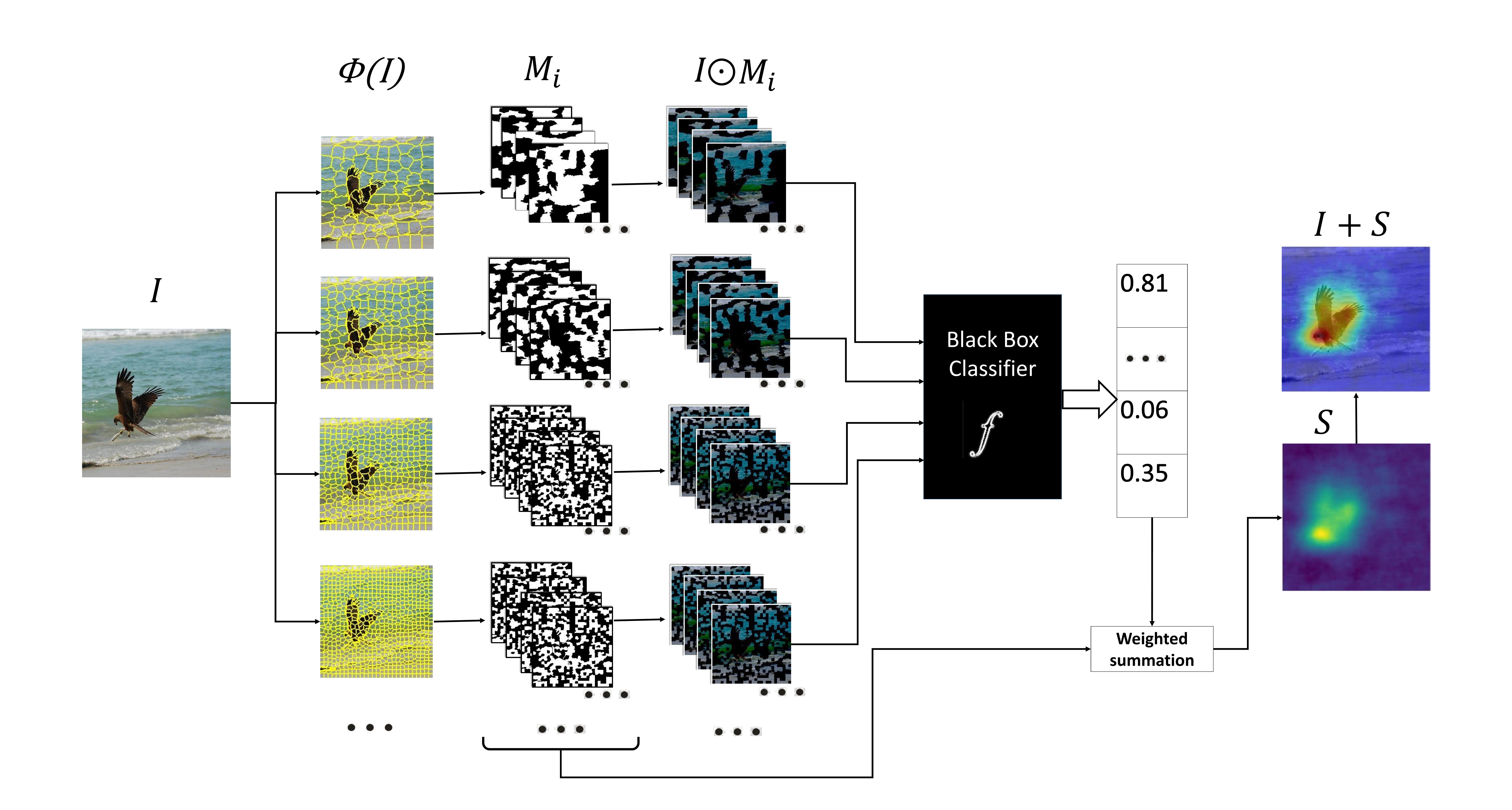}
	\end{center}
	\caption{Overview of MFPP method work flow: The input image $I$ is sent to a segmentation algorithm $\Psi$ to generate multiple-scale segmentation results. For each segmentation scale, masks $Mi$ are randomly generated and element-wise multiplied with input $I$ to obtain masked images. These masked images are fed to the black-box DNN model to obtain the saliency scores of the target classes. The scores are weighted summed across all masks. The final output is the saliency map.}
\label{fig:arch}
\end{figure*}

\subsection{Model-dependent vs. Model-agnostic}
Since the ultimate goal of model interpretation is to diagnose and analyze how the model works, the model itself is the protagonist. Whether the model under the interpretation is a black-box or a white-box divides different methods into two main camps: model-dependent methods (MDM) and model agnostic methods (MAM). MDM leverages internal model information to generate an explanation. For example, CAM generates importance map of the input image by using class-specific gradient information that flows into the final GAP (Global Average Pooling~\cite{gap}) layer of a CNN model. Grad-CAM requires of activation values and specified feature map. Grad-CAM++~\cite{gradcampp} even needs a smooth prediction function because it utilizes the third derivative. In short, model-dependent methods usually have strict restrictions on their use and often suffer from low-resolution results.

On the contrary, model-agnostic methods (such as LIME~\cite{lime}, BBMP~\cite{bbmp}, RISE~\cite{rise}, FGVis~\cite{fgvis} ) have no such restrictions, as they do not need to use model internal information. MAM only leverage input samples and corresponding output results to visually explain the model’s predictions. MAM can explain the predictions of almost all classifiers, including CNN models and classical machine learning models (\eg~support vector machine, decision tree, and random forests~\etal). Model-agnostic methods have broad practicality.

\subsection{Patch-wise vs. Pixel-wise perturbation}
Input perturbation is a general method, and we measure the output change after changing the input by removing or inserting information (such as masking, blurring, and replacing) from the image. It can also be used to generate volume prediction data for random sampling or local model fitting. BBMP evaluates and selects the most meaningful perturbation method by comparing area blur, fixed value replacement and noise added perturbation. The patch-wise and pixel-wise perturbation will lead to different visual results.
 
Pixel-wise perturbation is adopted by FGVis and real-time saliency~\cite{dabkowski2017real}, while LIME, Anchors~\cite{ribeiro2018anchors} and Regional~\cite{seo2018regional} are based on patch-wise perturbation. The perturbation method will affect the granularity of the output saliency map. In general, salience maps from pixel-wise methods are more accurate in terms of location, at the expense of lower semantics details because their results are spatially discrete. Results from patch-wise methods are more visually pleasing and close to model-dependent results where the boundaries can better fit the object boundaries.

\subsection{External model fitting vs. Statistical method}
Depending on how to handle the model output from the perturbation input, the different methods are further divided into two camps and lead to different results and processing time.

LIME needs to locally fit another linear model with ridge regression to facilitate interpretation. Anchors~\cite{ribeiro2018anchors} method further expands this idea to locally fit decision trees in order to provide better interpretability. Both methods require manually selecting the number of iterations for local model fitting. Manual configuration does not guarantee local model fits well. Both BBMP and EP use gradient-based optimization methods to iteratively optimize the saliency map (Adam~\cite{adam}). Our experiment results show that the number of iterations will significantly affect the accuracy and appearance of the final saliency map, however, the author sets the iteration number to a constant value based on experience.
In contrast, RISE uses statistical methods, which perform weighted summation of all sampled outputs without the need of additional model fitting or optimizer iterations. As shown in Table~\ref{tab:time}, the statistical method requires less processing time.

\section{Proposed Method}
\label{method}
Input perturbation is commonly used in existing methods [LIME, Anchors, BBMP, RISE, FGVis, EP]. In RISE~\cite{rise}, Petsiuk~\etal proposed a random input sampling method. They randomly mask half of 7x7 grids and additionally apply a random transportation shift to generate masks. This method ignores the morphological characteristics of the object, making the visual interpretation results unsatisfactory, as shown in the Figure~\ref{fig:benchmark}. In LIME~\cite{lime}, coarse-grained superpixel perturbation makes their interpretation coarse-grained and low-precision, as shown in Figure~\ref{fig:benchmark}. In Sec.\ref{sec:31}, We show the importance of morphology to visual interpretation results and analyze the relationship between the two. In Sec.\ref{sec:32}, we introduce a new input perturbation method, bridging multi-scale morphological information with the input perturbation method. The granularity of the interpretation results has been greatly improved with proposed method. Sec.\ref{sec:33} shows the overall flow chart of our proposed method.
\begin{figure}[ht]
	\centering
	\includegraphics[height=5.5cm,width=8cm]{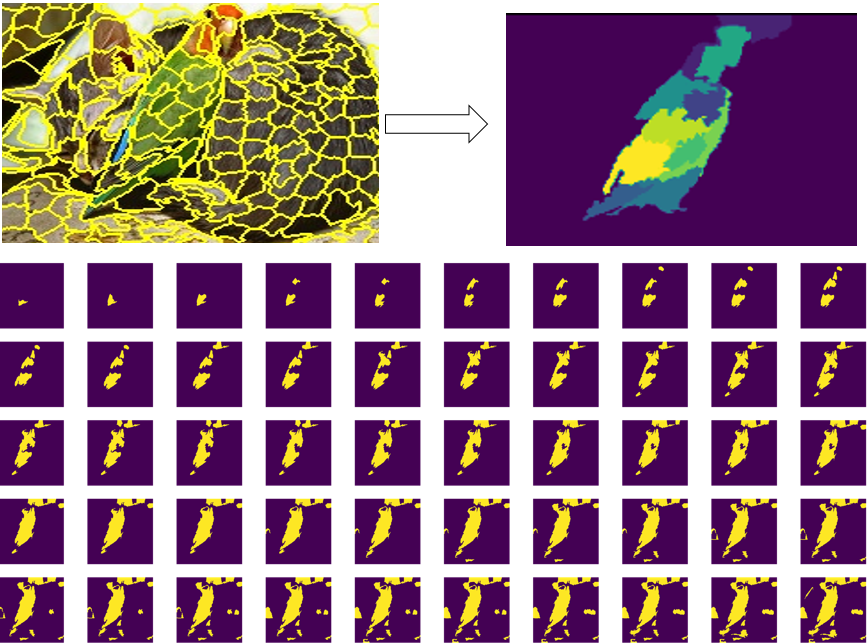}
	\caption{Top-left: the superpixel result for 'bird-and-cat' picture. Top-right: disturb superpixels and statistically measure the "bird" score of each segment; Bottom: visualize the probability distribution of bird fragments from high to low. }
\label{fig:patch}
\end{figure}

\subsection{Morphological analysis on visual explanation}
Objects and their surrounding areas are the main basis for the classification model to make decisions. For each prediction, the content composed of objects is very important for the visual interpretation of the results. According to morphological theory, objects consist of shapes, textures, and colors. Zeiler~\etal  ~\cite{zeiler2014visualizing} designed DeconvNet to visualize and understand convolutional networks. Their results show that the lower layer can understand edges, corners and colors. The middle layer focuses on the texture and part of the shape. Finally, the high layer will recognize the entire object and tend to extract the semantic information of the complete shape. Inspired by this work, we proposed the design to make full use of morphological information when performing input interference.

We started with an experiment. As shown in Figure~\ref{fig:patch}, first, the image is divided into different segments, also known as super-pixels. Then, we perturb the super-pixels and send perturbed image to the model $f$ for prediction. Third, we statistically measure the significance score of the "bird" category in each prediction. Finally, we lower the threshold of the interpretation score to visualize the probability distribution of the segments that make up the target object. 

Based on this method, we can evaluate the contribution of each patch of a specific category. For the "bird" type prediction, as shown in the upper right of Figure~\ref{fig:patch}, the fragments of different score are shown with different colors. The brighter the color, the higher the score. It can be seen that the high-scoring patches clustered where the bird is. As shown in the Figure~\ref{fig:patch}, the score gradually decreases from the center of the bird.

\begin{figure}[ht]
	\centering
	\includegraphics[height=3cm,width=8cm]{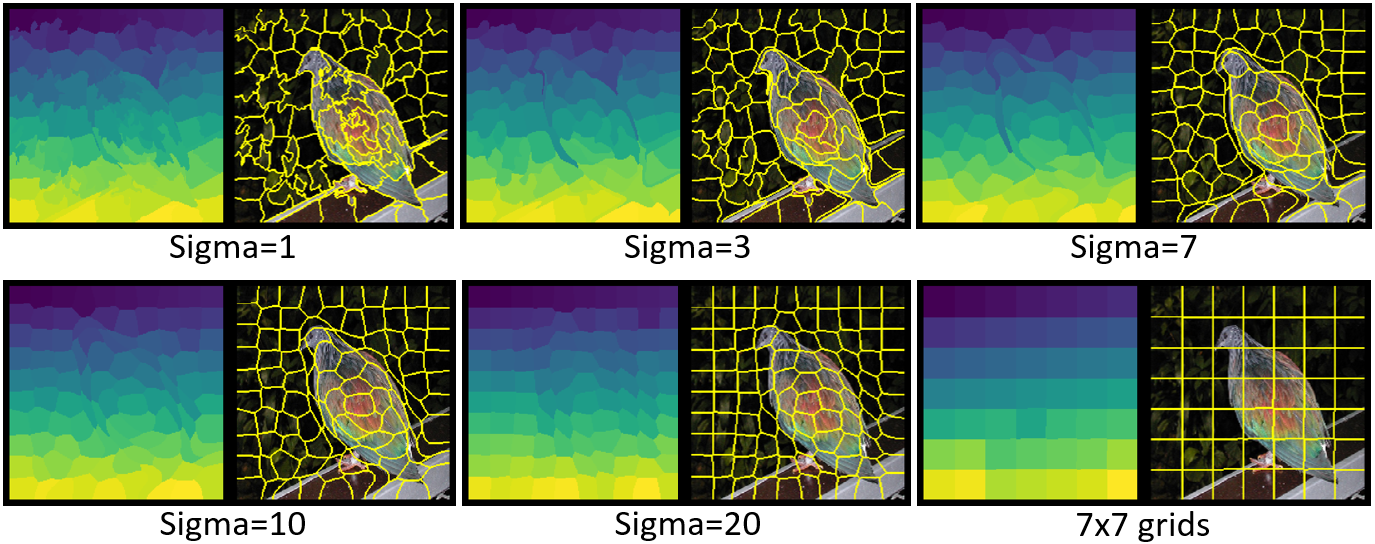}
	\caption{Different boundary areas generated using edge-based and grid-based boundaries. Top: Edge-based boundary. From left to right, sigma is 1, 3, 7; Bottom: Grid-based boundary. From left to right, sigma is 10 and 20. The bottom right corner is the 7x7 grid boundary used by RISE.}
\label{fig:seg}

\end{figure}

This observation inspired us to utilize morphological-based masks to better perturb the model input with its semantic distribution. Since segmentation is the basis for generating masks in our method, a fast and effective super-pixel algorithm SLIC~\cite{slic} with $O(N)$ time complexity is adopted. In Figure~\ref{fig:seg}, different sigma values are used in SLIC to generate different segmentation results. These sigma values control the smoothness of the segment boundaries, \ie morphological degree, it influences the explanation performance as shown in the sub-figure (a) of Figure~\ref{4charts}.

\label{sec:31}
\subsection{Fragments Perturbation Pyramid}
\begin{figure}[ht]
	\centering
	\includegraphics[height=3cm,width=8cm]{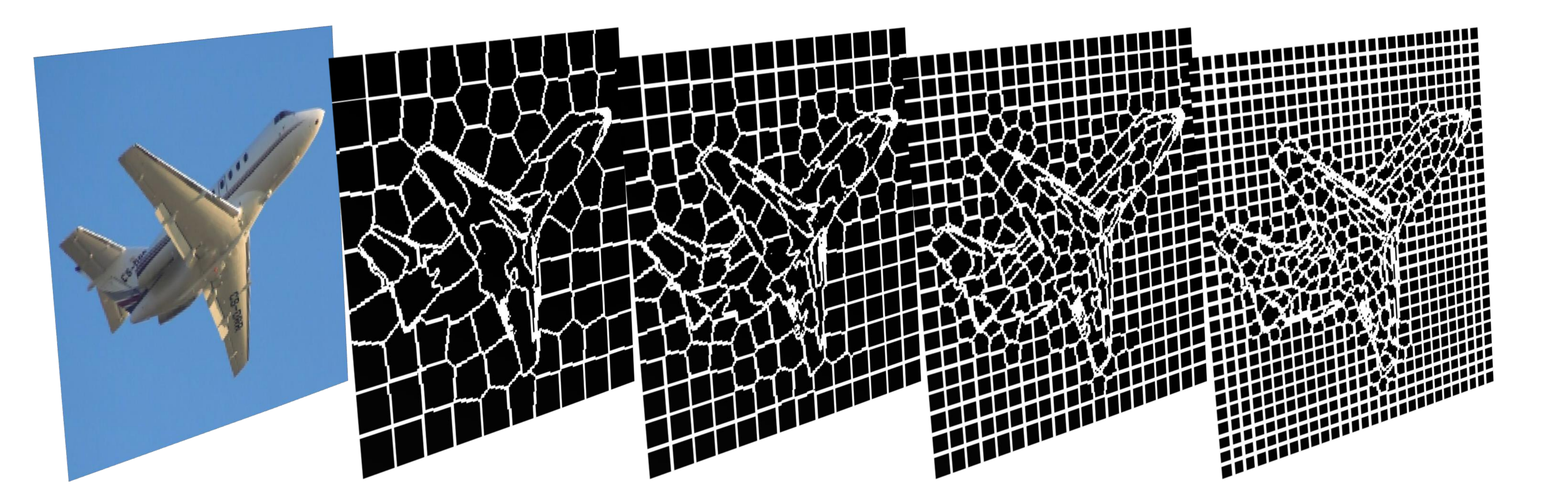}
	\caption{Different segmentation densities/scales are applied to the input image.}
\label{fig:scale}
\end{figure}

Lin~\etal~proposed Feature Pyramid Network (FPN)~\cite{fpn} to take advantage of the inherent multi-scale pyramid hierarchy of deep convolutional networks, which shows significant improvements in certain applications. In the field of object detection, from sliding windows to multiple feature maps as input to the classification module, people are always looking for a way to enable the predictor to detect and recognize objects of varying sizes.  In Faster-RCNN~\cite{fasterrcnn}, anchors are designed to expand the receiving field to different scales, thereby significantly improving the accuracy of objects of different scales. In addition, YOLO~\cite{yolo,yolov2,yolo3} series and other one-shot object detection networks continue to use different scales of feature maps in the evaluation phase. Based on this observation, we transfer this classic idea from object detection to model explanation. 

As shown in Figure~\ref{fig:scale}, by changing the segment number value in the segmentation algorithm, the input image $I$ is fed into the segmentation algorithm and is segmented into segments with different granularity. These fragments are used for randomized mask generation.

This method allows those model interpreters to view multi-scale objects in the input image, as shown in Figure~\ref {fig:scale}. This is helpful for enriching explanation sources.
\label{sec:32}
\subsection{MFPP}
Figure~\ref{fig:arch} shows the overall structure of MFPP and all data flows used to explain the prediction of a black-box model.

The input image $I$ is sent to the segmentation algorithm $\Psi$ to generate multiple segmented segments. Masks $M_{i}$ are generated by converting randomly selected fragments into zero grayscale. We multiply these masks with input $I$ element-wisely to get the masked images. Then the masked images are fed to the black-box model $f$ to get prediction score of target class, which is weighted summed to generate the final saliency map.

Let $I:\Omega\to\mathbb{R}^3$ be the space of a color image with size $H\times W$ where $\Omega=\lbrace1, . . . ,H\rbrace \times \lbrace1, . . . ,W\rbrace$ is a discrete lattice. Let $\Phi:\Gamma\to\mathbb{R}$ be a black machine learning model, which maps the image to a scalar output value $\Phi(I)$ in $\mathbb{R}$. The output could be activation, or class prediction score of$\Phi:\Gamma\to\mathbb{R}$.

Next, we investigate what part of the input
$I$ strongly activates the category, resulting in a large response $\Phi(I)$. In particular, we would like to find a mask $M$ and assign a value $M(\mu) \in \lbrace0, 1\rbrace$ to each pixel $\mu\in\Omega$, where $M(\mu) = 1$ means that the pixel contributes a lot to the output, and $M(\mu) = 0$ does not.
Based on Monte Carlo method, we can get 

\begin{eqnarray}\label{equ:S}
S_{I,\Phi}(\mu)\approx\frac{1}{E[M]\cdot N}\sum_{i=1}^{N}\Phi(I\odot M_{i})\cdot M_{i}(\mu)
\end{eqnarray}
where $\Psi$ is image segmentation operation, and $F_{l}$ is the output of image $I$ from this operation: 

\begin{eqnarray}
F_{l}=\Psi(I) 
\end{eqnarray}

In this case, $N$ is the total number of masks with different segmentation scales, $g(l)$ is the number of fragments in group $l$ and $L$ is the total number of groups.

\begin{equation}\label{equ:N}
N=\sum_{l=1}^{L}g(F_{l})
\end{equation}
Substituting N from (\ref{equ:N}) in (\ref{equ:S}) 
\begin{equation}
\resizebox{.91\linewidth}{!}{$
    \displaystyle
    S_{I,\Phi}(\mu)=\frac{1}{E[M]\cdot\sum_{l=1}^{L}g(F_{l})}\sum_{l=1}^{L}\sum_{i=1}^{g(F_{l})}\Phi(I\odot M_{l,i})\cdot M_{l,i}(\mu)
$}
\end{equation}

Please note that our method does not use any internal information in the model, so it is suitable for interpreting any black box model.

\label{sec:33}

\section{Experiments and Results}
\label{experiment}
\subsection{Datasets and based models}
Our algorithm is implemented and evaluated with PyTorch 1.2.0. The experiments are run with a single Nvidia P6000 GPU.  The the size of input image $I$ to black box model is 224x224.

In the intuitive experiment, the four SOTA methods and our proposed MFPP method were visually compared on the typical samples of MS-COCO 2014 dataset~\cite {Lin_2014}. Quantitative experiments were conducted on the entire test subset of the PASCAL VOC~\cite{everingham2010pascal} and MS-COCO 2014 minival set. The models used in the experiment are VGG16~\cite {simonyan2014deep} and ResNet50~\cite {He_2016}.

\subsection{Intuitive results}
In Figure~\ref {fig:benchmark}, we compare the model explanation result of the total $ 6 $ methods for same input images (the first column), including the proposed MFPP and its fast version. For intuitive visual evaluation, MFPP provides a more accurate and fine-grained saliency map than other competitive methods.

Grad-CAM's outputs look coarse with low resolution due to the upscaling operation in the saliency map generation. As shown in Figure \ref{fig:benchmark}, Grad-CAM can locate objects such as bananas in the first row and skiers in the fourth row, however the area it covers is too large and far beyond the object area borders, which reduces reliability. 
The weakness of Grad-CAM is clearly shown in the goldfish example (the fifth row), where there are multiple tiny objects of the same category.

Some existing model-agnostic methods (such as LIME, BBMP, RISE and FGVis) have accuracy and/or granularity issues. For example, the BBMP column shows that it has a poor explanation of the first $3$ samples, because when the background is complex and deceptive, BBMP cannot correctly locate the target.

For first $2$ samples of EP's output (7th column), its saliency map can only locate several points of the target, and it totally fails in the third example due to complex background of baseball player. 
It has similar weakness at handling  multiple tiny objects of the same category since it easily loses focus on non-extreme points, as shown in goldfish sample.

RISE can generate a pixel-wise heat-map, but its grid-based sampling method leads to two problems. First, the performance of non-rectangular shaped targets is poor. Second, the results are coarse granularity as shown in the baseball player sample (the third row).

The second and third columns show the performance of the proposed Fast-MFPP ($4000$ masks) and MFPP ($20000$ masks). In the first two and last rows, the proposed methods outperform all other methods. In the goldfish case, it finds most instances correctly without error. In the baseball case, the proposed designs are robust even with complex background, and are able to identify the exact body shape area of the target object.

For overall comparison, MFPP ($20000$ masks) generates the most accurate and cleanest importance map. The importance map is clear and pleasing, with minimal errors and noise. 

\subsection{Ablation Studies}
\label{ablation study}
In this section, we perform ablation studies to study the effects of various parameters used in our model. Pointing game accuracy conducted on ResNet50 and VOC07 test data set are used as performance metrics. 

\begin{figure*}[ht]
	\begin{center}
		\centering
		\includegraphics[height=4.3cm,width=17cm]{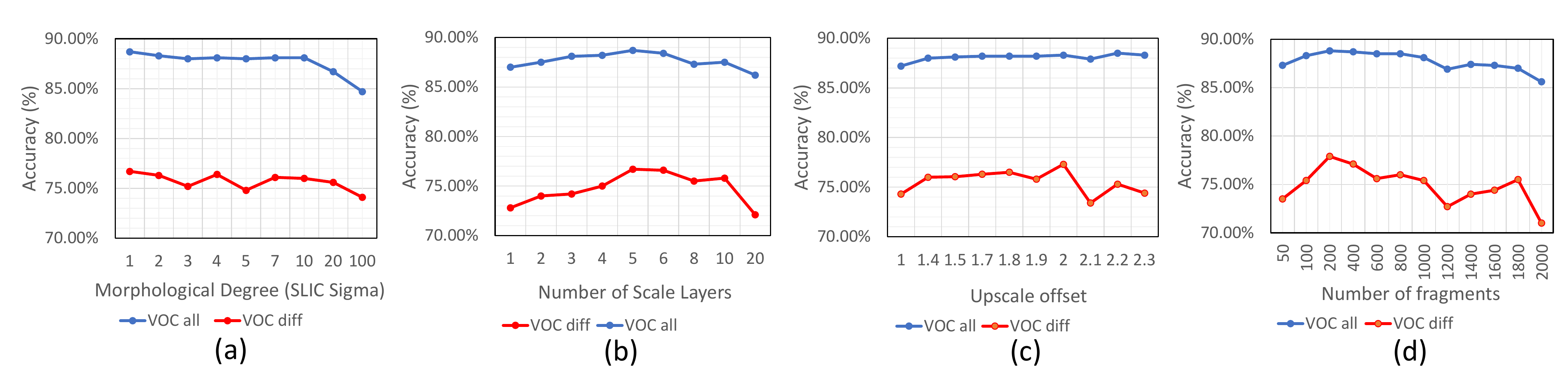}
	\end{center}
	\caption{Pointing game accuracy varies with different MFPP configurations: (a) shows the effect of morphological degree(\ie sigma value in SLIC, the smaller, the more morphological); (b) shows the effect of pyramid layers (number of fragmentation scales); (c) shows the effect of mask upscaling offset (masks are firstly generated in upscaled size and then random crop to target size); (d) shows the effect of fragment numbers in the case of one pyramid layer. All experiments are conducted on ResNet50 and VOC07 test set. }
\label{4charts}
\end{figure*}

Figure~\ref{4charts} sub-figure (a) depicts the sensitivity of MFPP to fragmentation morphological degree (the sigma value of SLIC algorithm). The higher the sigma value, the smoother the boundary. When sigma is less than $10$, MFPP is not very sensitive to it. When sigma is high enough, MFPP will be simplified to multi-layer RISE. Same effect is also shown in the lower right sub-picture of Figure~\ref{fig:seg}. Figure~\ref{4charts} sub-figure (b) shows the effect of the pyramid layer (\ie the number of fragmentation scales). In the beginning, more layers can improve performance. When the number of layers becomes too high, performance degrades. During mask generation, masks are firstly generated in upscaled size, and then randomly cropped to the target size. Figure~\ref{4charts} sub-figure (c) shows the sensitivity to mask upscaling offset. Figure~\ref{4charts} sub-figure (d) shows the impact of the number of fragments, and very dense or sparse fragments will reduce the final performance. 

For experiments in Figure~\ref{fig:benchmark} and the rest of experiments, the configurations of MFPP and Fast-MFPP are as follows: SLIC sigma is 1 and compactness is 10, upscale offset is 2.2, the number of fragments for 5 pyramid layers are $[50, 100, 200, 400, 800]$, respectively.

\subsection{Main results}
In this section, we quantitatively evaluate the performance of model explanation and processing time on the pointing game experiments~\cite{zhang2018top}. Pointing game experiment extracts maximum points from saliency map and measuring whether it falls into ground truth bounding boxes. The accuracy score is defined as $Acc=\frac{Hits}{Hits+Misses}$. The experiment was conducted on the PASCAL VOC07 test data set (with 4,952 images and 20 categories of ground truth labels) and the COCO 2014 minival data set (excluded images with the attribute "iscrowd = 1"). We repeated the experiment 3 times and took the average. The experiment results are compared to some reference method results extracted from \cite{ep}.

\begin{table}[!t]
\renewcommand{\arraystretch}{1.3}
\caption{The Result of Pointing Game~\cite{zhang2018top} on VOC2007 \textit{test} and COCO2014 \textit{minival} Dataset. Methods in Grey Color are for Black-box Model. EP's result on VOC07 is taken from~\cite{ep}.}
\centering
\begin{tabular}{|c|cc|cc|}
\hline
&\multicolumn{2}{c|}{VOC07 Test}&\multicolumn{2}{c|}{COCO14 MiniVal}\\
\cline{2-5}
Method & VGG16 & ResNet50& VGG16 & ResNet50\\
\hline
Cntr.& $69.6$ & $69.6$&27.6&27.6\\  
Grad & $76.3$ & $72.3$&37.4&35.4\\
DConv & $67.5$ & $68.6$&30.5&30.2\\ 
Guid. & $75.9$ & $77.2$&38.4&41.4\\ 
MWP & $77.1$ & $84.4$&39.2&48.8\\
cMWP & $79.9$ & $90.7$&49.8&57.4\\
Grad-CAM& $86.6$ & $90.4$&54.0&57.0\\
\hline
\rowcolor[gray]{0.8}
RISE & $86.4\pm0.6$ & $86.6\pm1.0$&$51.1\pm0.1$&$54.4\pm0.4$\\
\rowcolor[gray]{0.8}
EP & $\textbf{88.0}$ & $88.9$&$51.5\pm0.1$&$56.1\pm0.2$\\
\hline
\rowcolor[gray]{0.8}
Fast-MFPP& $86.1\pm0.2$ & $88.7\pm0.4$&$50.6\pm0.2$&$54.5\pm0.3$\\
\rowcolor[gray]{0.8}
MFPP & $87.0\pm0.1$ & $\textbf{89.1}\pm\textbf{0.6}$&$\textbf{52.0}\pm\textbf{0.2}$&$\textbf{56.4}\pm\textbf{0.2}$\\
\hline
\end{tabular}
\label{tab:game}
\end{table}
\begin{table}
\caption{The Benchmark for Average Processing Time for Single Sample Explanation on VOC07 Test Dataset.}
\centering
\begin{tabular}{|c|cc|cc|}
\hline
&\multicolumn{2}{c|}{VOC07 Test}&\multicolumn{2}{c|}{COCO14 MiniVal}\\
\cline{2-5}
Method & VGG16 & ResNet50& VGG16 & ResNet50\\
\hline
\rowcolor[gray]{0.7}
LIME & $39.8\pm0.9$ & $32.3\pm1.1$&$35.6\pm1.0$&$30.2\pm1.4$\\
\rowcolor[gray]{0.8}
RISE & $17.0\pm0.1$ & $13.5\pm0.1$&$\textbf{9.1}\pm\textbf{0.3}$&$19.8\pm0.4$\\
\rowcolor[gray]{0.8}
EP & $123.9\pm0.2$ & $72.1\pm0.1$&$92.6\pm3.5$&$75.9\pm0.5$\\
\hline
\rowcolor[gray]{0.8}
Fast-MFPP & $\textbf{16.9}\pm\textbf{0.1}$ & $\textbf{6.7}\pm\textbf{0.0}$&$\textbf{9.1}\pm\textbf{0.3}$&$\textbf{10.0}\pm\textbf{0.3}$\\
\rowcolor[gray]{0.8}
MFPP & $83.9\pm0.1$ & $32.6\pm0.1$&$45.1\pm0.3$&$50.1\pm0.2$\\
\hline
\end{tabular}
\label{tab:time}
\end{table}

The evaluation includes two versions of MFPP, which have different numbers of masks; Fast-MFPP contains $4000$ masks, and MFPP contains $20000$ masks. Both of them use the perturbation pyramid of $5$ layers. EP is the existing SOTA with multiple hyper-parameters, and we configure it with default values as follows: area is $0.1$, the maximum number of iterations is $800$, step is $7$, sigma is $21$, smooth is $0$. The configurations of RISE are as follows: $4000$ masks for VGG16 and $8000$ masks for ResNet50 to match RISE paper, the number of grid cells is 7x7, and the probability of closing is $0.5$.

As shown in Table~\ref{tab:game}, in terms of localization accuracy, for VOC07 test set, MFPP has the highest score on ResNet50(\textbf {89.1\%}), which is higher than current SOTA method EP~\cite{ep} (88.9\%). EP has the best accuracy on VGG16, which is 88.0\%. For COCO14 minival set, MFPP has the highest scores on both VGG16(\textbf{52.0\%}) and ResNet50(\textbf{56.4\%}). They are also higher than EP(51.5\% and 56.1\%)
Methods in grey rows applies to black-box models, while methods in white rows applies to white-box models. 
Overall, the two versions of MFPP are more accurate than most of existing methods.

For the processing time benchmark, we measure the total execution time from input image preprocessing to saliency map generation. We report both average and standard deviation of the total execution time. The hardware used during test is a single Nvidia P6000 GPU. As shown in Table~\ref{tab:time} , for VOC07 test set, MFPP takes 32.63 seconds per explanation on ResNet50, which is 2.2 times faster than EP's 72.09 seconds. It's also 1.47 times faster on VGG16. In particular, the fast version of Fast-MFPP is 10.7 times faster than EP on ResNet50 and 7.3 times faster on VGG16. The experiments on COCO14 minival set show similar results. The Fast-MFPP is the fastest methods compared to other other explaining black box models in both VOC07 test and COCO14 minival set.

\begin{figure}[ht]
	\centering
	\includegraphics[height=10.5cm,width=8.4cm]{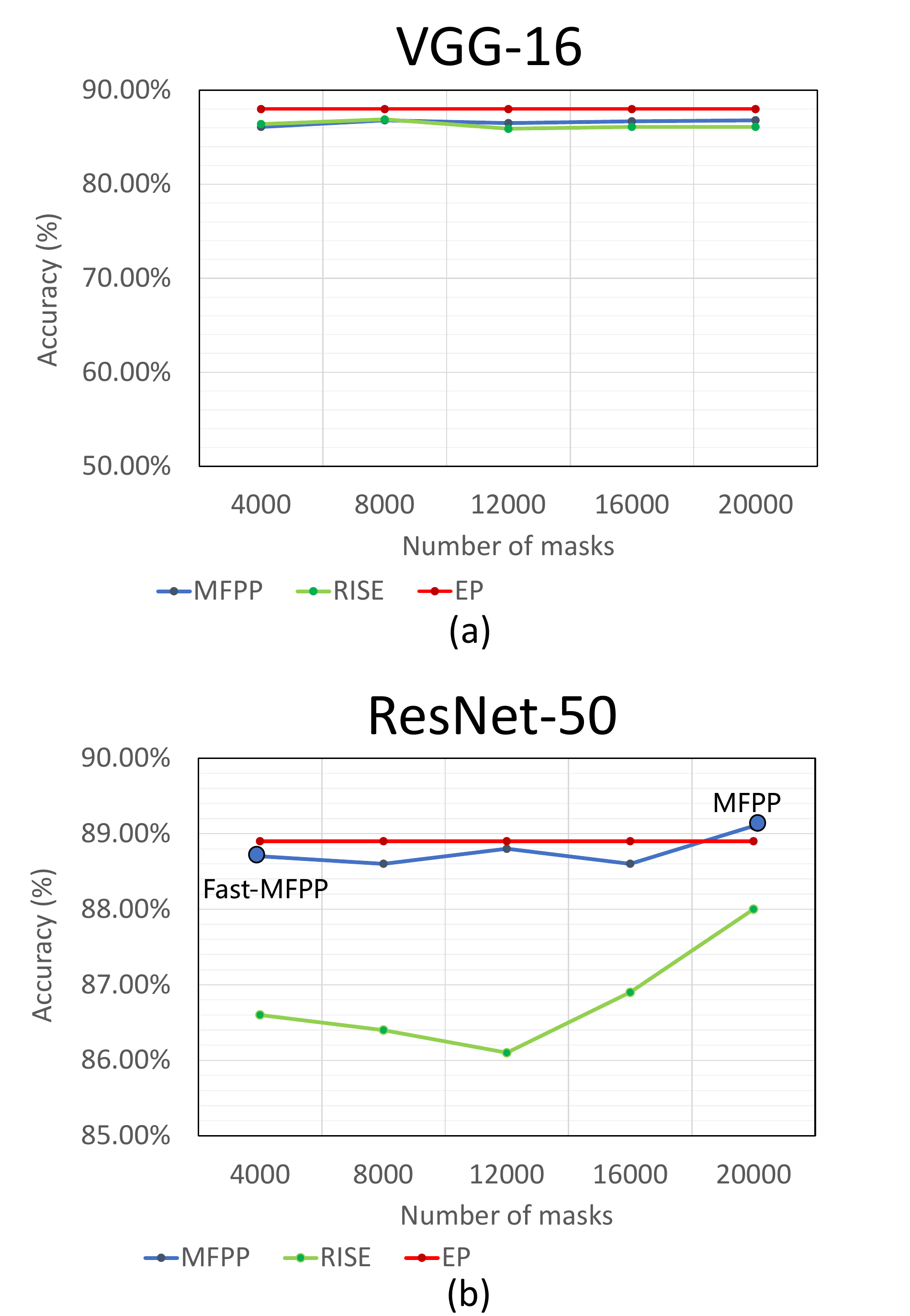}
	\caption{The accuracy of the pointing game experiment with different model explanation methods: (a) accuracy on VGG16 (b) accuracy on ResNet50. Fixed value for EP, since number of masks is not its parameter.}
\label{2charts_a}
\end{figure}

\begin{figure}[ht]
	\centering
	\includegraphics[height=10.5cm,width=8.4cm]{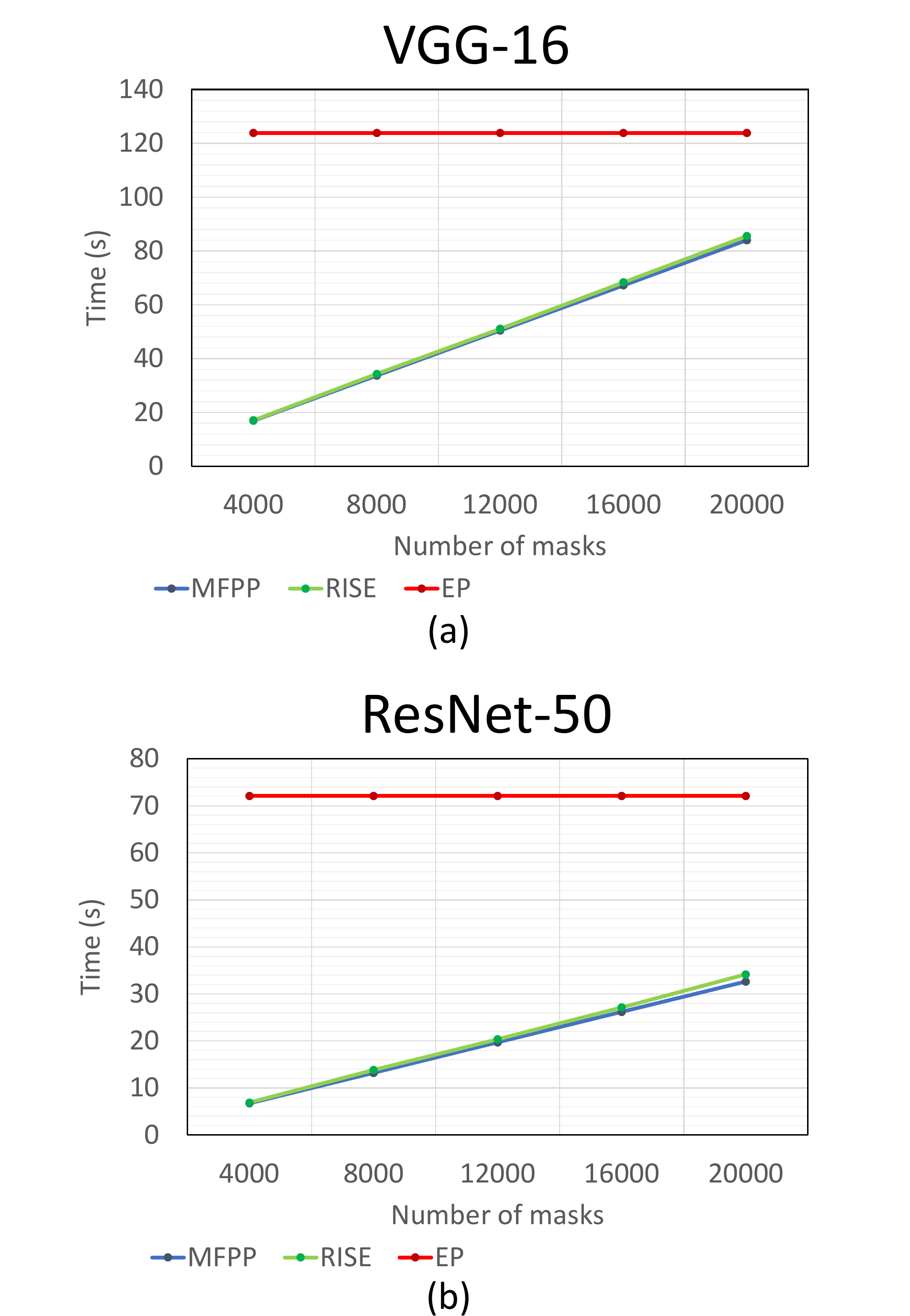}
	\caption{The processing time of the pointing game experiment with different model explanation methods: (a) processing time on VGG16 (b) processing time on ResNet50. Fixed value for EP, since number of masks is not its parameter.}
\label{2charts_t}
\end{figure}

In Figure~\ref{2charts_a}, (a) and (b) show the impact of mask number on the pointing game accuracy on VGG16 and ResNet50. The accuracy increases slowly with the number of masks increases. When MFPP uses $20000$ mask for ResNet50, it exceeds SOTA and achieves the highest score.

In Figure~\ref{2charts_t}, (a) and (b) show the corresponding processing times when using different mask numbers. The processing time of both MFPP and RISE increase linearly as the number of masks increases. They both are faster than EP.

In summary, pointing game experiments conducted on the PASCAL VOC07 test set and COCO minival set show that our proposed MFPP method benefits from morphological fragmentation and multiple perturbation layers. In terms of accuracy, it meets or exceeds the performance of the existing SOTA black box interpretation methods. At the same time, for the average processing time of each sample, MFPP is at least twice faster than the existing SOTA method. In particular, Fast-MFPP is more than 10 times faster than the existing SOTA method on ResNet50, and has the same accuracy level (within 0.5\%).
\label{evaluation}

\section{Conclusion}
\label{conclusion}
This paper proposes a novel MFPP method to explain the prediction of black box models with multi-scale morphological fragment perturbation modules. First, we prove that morphological fragmentation is a more effective perturbation method than random input sampling in model interpretation tasks. Secondly, MFPP can generate a finer-grained interpretation of critically shaped objects on an intuitive basis. Third, the quantitative experiments show that the performance of MFPP achieves or exceeds the SOTA black box interpretation method EP~\cite{ep} of the classic pointing game accuracy score. MFPP is at least twice as fast as EP. In particular, the fast version of MFPP is an order of magnitude faster than EP on ResNet50, while achieving the same level of accuracy. Since MFPP outperforms the latest methods in terms of accuracy and speed, we believe that MFPP can be a promising interpretation method in the field of deep neural network diagnosis.




\newpage

\bibliographystyle{IEEEtran}
\bibliography{MFPP}




\end{document}